\pdfoutput=1

\documentclass[11pt]{article}

\usepackage{emnlp2021}

\usepackage{times}
\usepackage{latexsym}

\usepackage{graphicx}
\usepackage{amssymb}
\usepackage{mathtools}
\usepackage{booktabs, multirow} 

\usepackage[T1]{fontenc}

\usepackage[utf8]{inputenc}

\usepackage{microtype}

\usepackage{inconsolata}
\title{Knowledge Injected Prompt Based Fine-tuning for Multi-label Few-shot ICD Coding}

\author{
Zhichao Yang$^{1}$, \ Shufan Wang$^{1}$, \ Bhanu Pratap Singh Rawat$^{1}$, \ Avijit Mitra$^{1}$, \ Hong Yu$^{1, 2}$ \\
$^{1}$ College of Information and Computer Sciences, University of Massachusetts Amherst\\
$^{2}$ Department of Computer Science, University of Massachusetts Lowell\\
{\tt \{zhichaoyang,shufanwang,brawat,avijitmitra\}@umass.edu } {\tt hong\_yu@uml.edu } \\
}

\begin{document}
\maketitle
\begin{abstract}
Automatic International Classification of Diseases (ICD) coding aims to assign multiple ICD codes to a medical note with average length of 3,000+ tokens. This task is challenging due to a high-dimensional space of multi-label assignment (tens of thousands of ICD codes) and the long-tail challenge: only a few codes (common diseases) are frequently assigned while most codes (rare diseases) are infrequently assigned. This study addresses the long-tail challenge by adapting a prompt-based fine-tuning technique with label semantics, which has been shown to be effective under few-shot setting. To further enhance the performance in medical domain, we propose a knowledge-enhanced longformer by injecting three domain-specific knowledge:  hierarchy, synonym, and abbreviation with additional pretraining using contrastive learning. Experiments on MIMIC-III-full, a benchmark dataset of code assignment, show that our proposed method outperforms previous state-of-the-art method in 14.5\% in marco F1 (from 10.3 to 11.8, P<0.001). To further test our model on few-shot setting, we created a new rare diseases coding dataset, MIMIC-III-rare50, on which our model improves marco F1 from 17.1 to 30.4 and micro F1 from 17.2 to 32.6 compared to previous method.



\end{abstract}

\section{Introduction}
Multi-label learning has many real-word applications in natural language processing (NLP), including but not limited to academic paper labeling \citep{chen-etal-2020-hyperbolic}, news framing \citep{akyurek-etal-2020-multi}, waste crises response \citep{Yang2020RiskRF}, amazon product labeling \citep{McAuley2015InferringNO, Dahiya21b}, and medical coding \citep{Atutxa2019InterpretableDL}. 
In contrast to multi-class classification, an instance in multi-label learning is frequently linked with more than one class labels, making the task more challenging due to the combination of potential class labels. 

In real-world tasks, there are often insufficient training data for 
rare class labels. 
Taking automatic international classification of diseases (ICD) coding as example, 
given discharge summaries notes as input, the task is to assign multiple ICD disease and procedure label codes associated with each note. The assigned codes need to be accurate and complete for the billing purposes.
As an example, the MIMIC-III dataset \citep{Johnson2016MIMICIIIAF} contains 8,692 unique ICD-9 codes, among which 4,115 (47.3\%) codes occur less than 6 times and 203 (2.3\%) occur zero times. Clinical practice requires a high accuracy, hence, it is not acceptable 
for a multi-label classifier to fail a disease diagnosis (or code assignment) because it is rare, since such a diagnosis may be of the most clinical importance for the patient. Therefore, the classifier is required to perform with high precision even for infrequent codes. This translates to data sparsity due to availability of few training examples.

\begin{figure*}[t]
	\centering
	\includegraphics[width=0.99\textwidth]{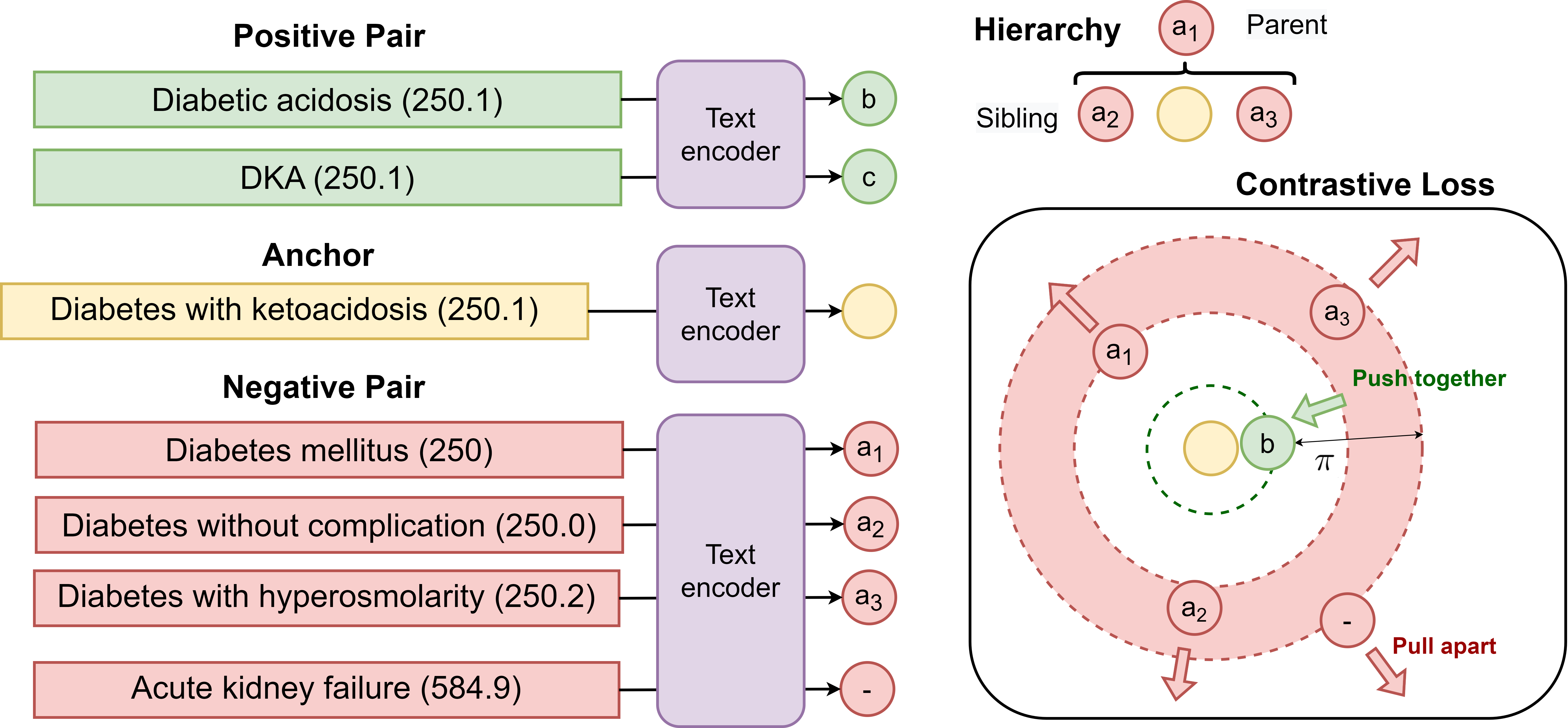}
	\caption{An illustration of self-alignment pretraining from medical knowledge UMLS, including the usage of (a) Hierarchy, (b) Synonym, (c) Abbreviation. Pink region is the dynamic margin ranges from $\pi/2$ to $\pi$ where we wish to pull negatives apart with a dynamic distance.}
	\label{fig:kpretrain}
\end{figure*}

To mitigate the data sparsity problem, additional structured knowledge could be applied. ICD codes are organized with an ontological/hierarchical structure where a text description is associated to each code. For instance, ICD 250 (Diabetes mellitus), shown in Figure \ref{fig:kpretrain}, is the parent of several child codes including 250.0 (Diabetes mellitus without mention of complication), 250.1 (Diabetes with ketoacidosis), and 250.2 (Diabetes with hyperosmolarity). Such child ICD codes are more semantically different from each other than their parent code 250. 

Synonyms including acronyms and abbreviations are common in medical notes. For instance, the description of code 250.00 is disease "type II diabetes mellitus". However, this code can be described in different text forms such as "insulin-resistant diabetes", "non-insulin dependent diabetes", "DM2", and "T2DM". Therefore, one naive way to assign ICD codes is to identify matching between candidate code descriptions and their synonyms in medical notes. In this work, we separate synonyms from both acronyms and abbreviations due to its importance in medical domain \cite{yu2002mapping}. While synonymous relations could be implicitly learned from pretrained language model (LM)
\citep{Michalopoulos2022ICDBigBirdAC, Li2022ClinicalLongformerAC}, previous researches show that language models are only limited biomedical \citep{sung-etal-2021-language} or clinical knowledge bases \citep{yao2022extracting} due to the data sparsity challenge in the medical domain. 
An explicit way of adding such medical knowledge into language model should be explored.

In this paper, we present a simple but effective Knowledge Enhanced PrompT (KEPT) framework. 
We implement and evaluate KEPT using a LM based on Longformer because clinical notes are typically more than 500 tokens. 
Specifically, we first $\mbox{pretrain}_{mimic}$
a Longformer LM on MIMIC-III dataset. 
Then, we further
$\mbox{pretrain}_{umls}$ 
on structured medical knowledge UMLS (Unified Medical Language System) using self-alignment learning with contrastive loss to inject medical knowledge into pretrained LM. 
For the downstream ICD-code assignment fine-tuning, we add a sequence of ICD code descriptions (label semantics) as prompts in addition to each clinical note as KEPT LM input. This allows early fusion of code descriptions and the input note. 
Experiments on full disease coding (MIMIC-III-full) and common disease coding (MIMIC-III-50) show that our KEPTLongformer outperforms previous SOTA MSMN \citep{Yuan2022CodeSD}. In order to test its few-shot ability, we create a new few-shot rare diseases coding dataset named MIMIC-III-rare50, and results show significant improvements compared between MSMN and our method. To facilitate future research, we publicly release the code and trained models\footnote{\url{https://github.com/whaleloops/KEPT}}.

\section{Related Work}

\subsection{Prompt-based Fine-tuning}
Prompt-based fine-tuning has been shown to be effective in few-shot tasks \citep{le-scao-rush-2021-many, gao-etal-2021-making}, even when the language model is relatively small \citep{schick-schutze-2021-just} because they introduce no new parameter during few shot fine-tuning. 
Additional tuning techniques such as to tune bias-term or language model head have shown to be efficient on memory and training time \citep{ben-zaken-etal-2022-bitfit, logan-iv-etal-2022-cutting}. 
However, most previous works focus injecting knowledge into prompt on single-label multi-class classification task \citep{hu-etal-2022-knowledgeable, wang-etal-2022-automatic, Ye2022OntologyenhancedPF}. To the best of our knowledge, this is the first work that applies prompting to multi-label classification task.

\subsection{Entity Representation Pretraining}
Many recent researches use synonyms to conduct biomedical entity representation learning \citep{sung-etal-2020-biomedical, liu-etal-2021-self, lai-etal-2021-bert-might, angell-etal-2021-clustering, Zhang2021KnowledgeRichSF, kong-etal-2021-zero, seneviratne-etal-2022-networks}. 
Our work is most similar to \citet{liu-etal-2021-self}, which uses additional pretraining scheme that self-aligns the representation space of biomedical entities from pretrained medical LM. They collect self-supervised synonym examples from the biomedical ontology UMLS, and use multi-similarity contrastive loss to keep the representation of similar entities closer to each other, before fine-tuning them to the downstream specific task.
However, their work differs from ours in (1) their testing being limited to only medical entity linking tasks and (2) not using hierarchical information, which has been shown to be useful in KRISSBERT \citep{Zhang2021KnowledgeRichSF}. In contrast to KRISSBERT, our contrastive learning selects negative samples from siblings (1-hop nodes) instead of random nodes in the graph. Our method follows InfoMin proposition that selected samples should contain as much task-relevant information while discarding as much irrelevant information in the input as possible \citep{Tian2020WhatMF}.

\subsection{ICD Coding}
ICD coding uses NLP models to predict expert labeled ICD codes given discharge summaries as input.
Currently, the most straightforward method is to take the best language model for encoding notes, and later use the label attention mechanism to attend labeled ICD codes to input notes for prediction \citep{Mullenbach2018ExplainablePO}. In comparison, we apply attention between codes and notes way before within the encoder with the help of prompt.
The label representations in attention played an important role in many previous works.
\citet{Li2020ICDCF} and \citet{Vu2020ALA} first randomly initialize the label representations. \citet{Chen2019AutomaticIC, Dong2021ExplainableAC, zhou-etal-2021-automatic} initialize the label representation with code description from shallow representation using Word2Vec  \citep{Mikolov2013EfficientEO}. \citet{Yuan2022CodeSD} further add description synonyms semantic information. 
In comparison, we use deep contextual representation from Longformer pretrained on both MIMIC and UMLS with contrastive loss. Similar pretrained language models have shown to be effective in previous works \citep{Wu2020CLEARCL, huang-etal-2022-plm, DeYoung2022EntityAI, Michalopoulos2022ICDBigBirdAC}. 

As stated previously, the high dimensions of available label codes, such as 14,000 diagnosis codes and 3,900 procedure codes in ICD-9 and 80,000 in industry coding \citep{Ziletti2022MedicalCW}, makes ICD coding challenging. Another challenge is the long-tail distribution, in which few codes are frequently used but most codes may only be used a few times due to the rareness of diseases \citep{Shi2017TowardsAI, Xie2019EHRCW}. 
\citet{Mottaghi2020MedicalSR} use active learning with extra human labeling to solve this issue.
Other recent works focus on using additional medical domain-specific knowledge to better understand the few training instances \citep{Cao2020HyperCoreHA, Song2020GeneralizedZT, Lu2020MultilabelFL, falis-etal-2022-horses, wang-etal-2022-novel}. 
\citet{Wu2017SemEHRAG} perform entity linking to identify medical phrase in document note.
\citet{Xie2019EHRCW} map label codes as entities in medical hierarchy graph. Compared to a baseline which uses a shallow convolutional neural network to learn n-gram features from notes, they add complex hierarchy structure between codes by allowing the loss to propagate through graph convolutional neural network. 
In contrast with the previous systems which adopt complex pipelines and different tools, our method applies a much simpler training procedure by incorporating knowledge into language model without requiring any knowledge pre or post-processing (i.e. MedSpacy, Gensim, NLTK) during the fine-tuning. 
Additionally, previous methods use knowledge graph as an input source, however, we train our language model to include knowledge graph as a target with contrastive loss.

\section{Methods}

\begin{figure*}[t]
	\centering
	\includegraphics[width=0.99\textwidth]{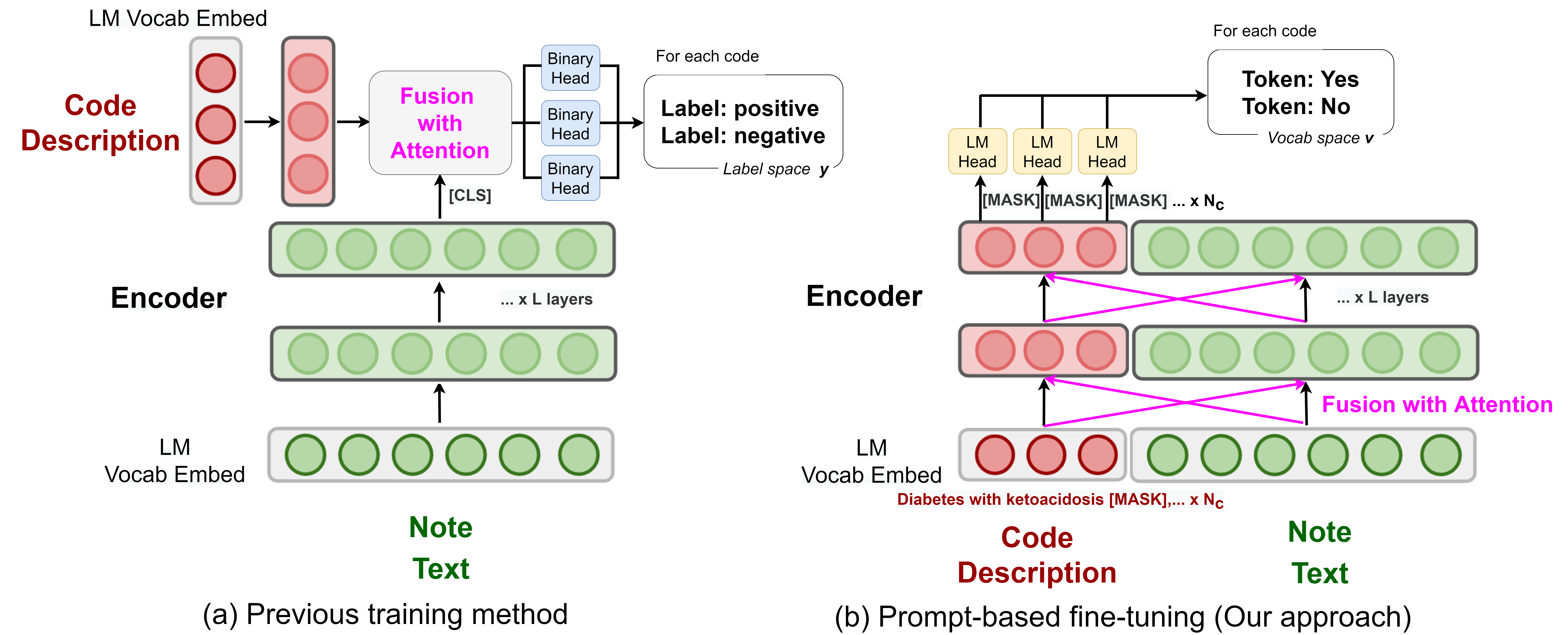}
	\caption{An illustration of (a) standard training method and (b) our proposed prompt-based fine-tuning.}
	\label{fig:model}
\end{figure*}

\textbf{ICD coding:}
ICD coding is 
a multi-label multi-class classification task.
Specifically, considering thousands of words from an input \textcolor{ForestGreen}{medical note} \textcolor{ForestGreen}{$t$}, the task is to assign a binary label $y_i \in \{0,1\}$ for each ICD code in the label space $Y$, where 1 means that note is positive for an ICD disease or procedure and $i \in$ range $[1,N_c]$. In this study, we define and evaluate the number of candidate codes $N_c$ as 50, although $N_c$ could be higher or lower depending on specific applications. 
Each candidate code has a short code description phrase $c_i$ in free text. For instance, code 250.1 has description \textit{diabetes with ketoacidosis}. \textcolor{BrickRed}{Code descriptions} \textcolor{BrickRed}{$c$} is the set of all $N_c$ number of $c_i$.

\subsection{Encoding Text with Longformer} \label{sec:enc}
To solve this task, we first need to encode free text into hidden representation with a pretrained clinical longformer. Specifically, we convert free text $a$ to a sequence of tokens $x_a$, the vocab embedding then maps $x_a$ to a sequence of hidden vectors. Next, the 1st layer of LM encoder attends one hidden vector to another hidden vector in the sequence with self-attention mechanism. This encoding process is repeated $l$ times to produce a sequence of final contextual hidden vectors $\pmb{h}_a\in \mathbb{R}^{L_t \times H_d}$ for each free text $a$ where $H_d$ is the hidden layer dimension and $L_t$ is the number of token in \textcolor{ForestGreen}{$t$}. 

\subsection{Fine-tuning with Prompt} \label{sec:prompt_method}

Prompt based fine-tuning is different from standard fine-tuning.
During standard fine-tuning, we usually make input $x_a =$ [CLS] a, where $a \in\{t,c_1, c_2, ..., c_{N_c}\}$. 
To assist LM in finding a mention of a label code in note text, we \textcolor{CarnationPink}{fusion} final contextual hidden representation of note text \textcolor{ForestGreen}{$t$} and code description \textcolor{BrickRed}{$c$} with attention.
Specifically, we first build code description representation \textcolor{BrickRed}{$\pmb{h'}_c$} $\in \mathbb{R}^{N_c \times H_d}$ by concatenating encoded hidden vector \textcolor{BrickRed}{$\pmb{h}_{c_i}\{\mbox{[CLS]}\}$}  $\in \mathbb{R}^{H_d}$ of token $\mbox{[CLS]}$ for each code description \textcolor{BrickRed}{$c_i$}.
We then build note aware code representation $\pmb{h}_f \in \mathbb{R}^{N_c \times H_d}$ for each code using cross attention between sequence of vectors \textcolor{BrickRed}{$\pmb{h'}_c$} as query and sequence of vectors \textcolor{ForestGreen}{$\pmb{h}_t$} as key, with attention weight $\alpha_{ij}$ between $i$th item in query and $j$th item in key as follow:

\vspace{4mm}
$\alpha_{ij} = softmax( (\pmb{W}_q \textcolor{BrickRed}{\pmb{h}_{c_i}\{\mbox{[CLS]}\}} ) (\pmb{W}_k \textcolor{ForestGreen}{\pmb{h}_{t_j}}) )$
\vspace{4mm}

\noindent where $\pmb{W}_q$ and $\pmb{W}_k$ are query weight and key weight to be trained.
To learn the probability of a code to assign, we train a binary label head, $softmax(\pmb{W}_b\pmb{h}_f)$, by maximizing log-probability of correct label for each code. An illustration of such a standard fine-tuning pipeline is provided in Figure \ref{fig:model} (a). This standard fine-tuning approach introduces many new parameter weights (589,824 with cross attention and 1,536 with binary label head for longformer), making it hard to learn in few shot setting where the number of training data is limited for each code \citep{gao-etal-2021-making}.
Similar training approaches were carried out in previous researches \citep{Mullenbach2018ExplainablePO, Li2020ICDCF, Kim2021ReadAA, luo-etal-2021-fusion, Sun2021MultitaskBA, zhou-etal-2021-automatic}
(specific label attention calculation may differ) 
, instead of a pretrained language model, they used unpretrained LSTM or CNN to encode free text, which added more untrained parameters during ICD training.

An alternative approach to multi-label classification is \textbf{prompt based fine-tuning}, where masks in prompt are filled-in by LM in cloze style \citep{gao-etal-2021-making}. We reformulate multi-label classification tasks with free text prompt template as input:

\vspace{3mm}
\footnotesize 
\hspace{-4mm}
$x_p =$\textcolor{BrickRed}{$c_1$} : [MASK] , \textcolor{BrickRed}{$c_2$} : [MASK] , ... ,  \textcolor{BrickRed}{$c_{N_c}$} : [MASK] .  \textcolor{ForestGreen}{t} . 

\vspace{3mm}
\normalsize

\noindent and use LM to decide if note is positive (or negative) for a code by filling [MASK] with vocab token yes (or no). This step is repeated $N_c$ times for each [MASK] and associated code \textcolor{BrickRed}{$c_i$}. 
Specifically, 
we encode free text prompt as mentioned before, and obtain final hidden vectors $\pmb{h}_p$ for input $x_p$. Notice that this encoding step would \textcolor{CarnationPink}{fusion} \textcolor{BrickRed}{code descriptions} and \textcolor{ForestGreen}{note text} with self-attention in every layer of LM encoder. 
We define a mapping function $M$ from $y_i$ in label space to vocab tokens as:

\begin{equation}
M(y_i) =
\begin{dcases*}
``yes"   & if $y_i =1$;\\
``no"    & if $y_i =0$; \\
\end{dcases*}
\label{eq:my}
\end{equation}

\noindent where $i \in$ range $[1, N_c]$. In this way, we transfer downstream multi-label classification task into a mask language model task like pretraining. For $i$th code, the label probability would be calculated as:

\small
\begin{equation}
\begin{aligned}
P(y_{i} | x_p) &= P(\mbox{[MASK]}_{c_i} = M(y_{i}) | x_p) \\
&= \frac{\mbox{exp}(\pmb{W}_{M(y_{i})} \cdot \pmb{h}_{p}\{\mbox{[MASK]}_{c_i}\})} {\sum_{\dot y \in Y} \mbox{exp}(\pmb{W}_{M(\dot y)} \cdot \pmb{h}_{p}\{\mbox{[MASK]}_{c_i}\})}
\end{aligned}
\label{eq:mPLM}
\end{equation}
\normalsize

\noindent where $h_{p}\{\mbox{[MASK]}_{c_i}\} \in \mathbb{R}^{H_d}$ is the hidden vector of the [MASK] associated with each code \textcolor{BrickRed}{$c_i$} in input $x_p$, and $W_M$ is the original parameter pretrained in LM head. Prompt based fine-tuning reuses all parameters during pretraining, and does not introduce new parameters, making the whole model easy to fine-tune in a few-shot setting. 

\subsection{Hierarchical Self-Alignment $\mbox{Pretrain}_{umls}$ (HSAP) using Knowledge Graph UMLS}
Since no new parameters are added to LM in prompt based fine-tuning, the performance on medical downstream task heavily relies on the quality of clinical pretrained LM. However, encoded hidden representations of similar medical terms are not guaranteed to be close to each other. Thus we apply self-alignment pretraining \citep{liu-etal-2021-self} to align similar terms closer to each other with additional knowledge. This additional $\mbox{pretrain}_{umls}$ is after masked language $\mbox{pretrain}_{mimic}$ and before auto ICD finetuning.
We first build self-supervised data from synonyms, abbreviations, hierarchy in the medical knowledge graph of the UMLS and ICD ontology (\textsection \ref{sec:sapa}), and inject such structural knowledge into a LM by pretraining it on self-supervised data with hierarchical contrastive loss (\textsection \ref{sec:sapb}).

\subsubsection{Generating Self-Supervised Data} \label{sec:sapa}
To generate pretraining examples, we first build a mapping between medical terms and codes as entities in the medical knowledge graph UMLS. Specifically, synonyms of an entity are collected from multiple English free text descriptions of entity via UMLS "MRCONSO" table. Abbreviations of an entity are collected from multiple English free text descriptions of entity in UMLS SPECIALIST Lexicon and Lexical Tools "lrabr" table. Medical terms of an entity is defined as the union of synonyms set and abbreviations set of an entity. To sample negative examples for contrastive loss, we then build a hierarchy tree of entities using ICD-9 code ontology. For example, ICD 250 (Diabetes mellitus) is the parent of ICD 250.0 (Diabetes mellitus without mention of complication), and ICD 250.1 (Diabetes with ketoacidosis) is the sibling of ICD 250.0 (Diabetes mellitus without mention of complication).

\subsubsection{Contrastive Learning} \label{sec:sapb}
Given self-supervised data, we further train clinical longformer using contrastive learning, with the intention of pushing target medical terms and positive medical terms closer, while pulling negative medical terms further away. We formulate this problem into hierarchical triplet loss on  a sampled mini batch encoded by LM.

\textbf{Encoding Medical Terms}:
Each medical term is usually a short phrase of multiple tokens. Similar to Phrase-BERT \citep{wang-etal-2021-phrase}, a medical term is encoded into a sequence of hidden vectors as descried in \textsection \ref{sec:enc}. We use clinical longformer \citep{Li2022ClinicalLongformerAC} as encoder for this process. We define a medical term's hidden representation $\pmb{p}$ as the first item in hidden vector sequence, which has shown to be effective in \citet{toshniwal-etal-2020-cross}.

\textbf{Hierachical Neighbor Sampling}:
We randomly select $i$ number of target anchor entities from the ICD hierarchy level $l$. Each medical terms represents a disease class. Collecting entities from each level could preserve the diversity of samples in the mini batch.
Then $j-1$ parents and siblings are randomly chosen for each of $i$ entities. The purpose of choosing \textbf{intra-class} parents and siblings is to encourage model to discriminate anchor entities from close neighbor entities. Finally, $k$ medical terms for each entity are randomly collected, resulting in $n = ijk$ medical terms in a mini batch $B$ of hierarchy level $l$. We collect mini batch from other hierarchy levels in the same way.

\textbf{Minibatch Triplet Loss with Dynamic Margin}:
Similar to \citet{Ge2018DeepML}, hierarchical triplet loss of a mini batch $B$ can be formulated as: 

\small
\begin{equation}
\begin{aligned}
L_{B} = \frac{1}{2N_B} \sum_{T_x \in T_B} \max(0, {m_x-| \pmb{p}^a_x-\pmb{p}^-_x| + |\pmb{p}^a_x-\pmb{p}^+_x|} )
\end{aligned}
\label{eq:contrast}
\end{equation}
\normalsize

\noindent where $T_B$ is all the triplets in the minibatch $B$. $N_B$ is the number of triplets in minibatch $B$, and each triplet $T_x$ consists of an anchor sample $\pmb{p}^a_x$,
a positive sample $\pmb{p}^+_x$ from positive class, a negative sample $\pmb{p}^-_x$ from intra-class or inter-class negative class. $m_x$ is a dynamic margin. It is computed according to entity's clinical term similarity between the anchor class entity and the negative class entity \citep{Zakharov20173DOI}. Specifically, for a triplet $T_x$, the dynamic margin $m_x$ is computed as:

\small
\begin{equation}
m_x =
\begin{dcases*}
\pi/2    & if \mbox{parent} ;\\
\pi/2 + arccos(|\pmb{p}^a_x \cdot \pmb{p}^-_x|)   & if \mbox{siblings} ;\\
\epsilon    & else ($\epsilon = \pi$). \\
\end{dcases*}
\end{equation}
\normalsize

\noindent where condition clause $\mbox{parent}$ and $\mbox{siblings}$ means that negative sample $\pmb{p}^-_x$ comes from intra-class parent and siblings of anchor sample. In practice, we set $\epsilon = \pi$. 
Thus, inter-class negative sample would be at least $\pi$ distance away from anchor sample, while intra-class negative sample would be at least $d \in [\pi/2, \pi]$ range distance away from anchor sample.
Such dynamic margin is different from constant margin in previous contrastive loss work in medical domain \citep{liu-etal-2021-self, Zhang2021KnowledgeRichSF}, and has shown to be effective in visual retrieval task in computer vision \citep{Ge2018DeepML}.


By minimizing loss defined in Equation \ref{eq:contrast}, we pretrain a medical knowledge injected clinical longformer. We then use such longformer to encode prompt and context (\textsection \ref{sec:prompt_method}), and thus gain knowledge injected prompt for downstream coding task. 

When applied to MIMIC-III-full data, it is infeasible to encode all 8,692 candidate ICD codes in the prompt
due to high memory cost (to be specified in \textsection \ref{sec:Limitations}). Instead, we used a two-stages approach. Specifically, we used model MSMN as 1st stage coder to select top 300 candidate codes, and then use our KEPTLongformer as 2nd stage coder to further narrow down the candidates to final prediction. Our 2nd stage coder functions similar to reranker in passage ranking \citep{Nogueira2019PassageRW}.

\section{Experiments}

\subsection{Dataset}

MIMIC-III dataset \citep{Johnson2016MIMICIIIAF} contains data instances of de-identified discharge summaries with expert labeled ICD-9 codes. The discharge summaries are from real patients. We applied the following text pre-processing step before tokenizer: (1) removing all de-identification tokens; (2) replacing characters other than punctuation marks and alphanumerical into white space (e.g. /n); (3) stripping extra white spaces. 
Previous work \citep{Mullenbach2018ExplainablePO} truncated discharge summaries at 4,000 words. Since longformer used tokens instead of words, we truncated discharge summaries at 8,192 tokens unless otherwise specified. This roughly aligns with our observation that word token ratio is about 1:2. Since procedure codes are related to subjective section of the note \citep{yang-yu-2020-generating}, we include relevant sections of discharge summaries for those length exceeds 8,192, and remove irrelevant sections such as discharge followup. The header names of the relevant sections are provided in Table \ref{tab:header}. We named this dataset \textbf{MIMIC-III-full}.

For the top-50 frequent codes prediction task, we filtered each instance that has at least one of the top 50 most frequent codes, and used the same splits as the previous work \citep{Vu2020ALA, Yuan2022CodeSD}. We named this dataset \textbf{MIMIC-III-50}. Detailed statistics are included in Table \ref{tab:data}.

To benchmark auto ICD coding task on few-shot learning, we also created a rare-50 codes prediction using  original MIMIC-III dataset. Among 8,692 different types of ICD-9 codes, we first selected codes with less than 10 times occurrences to fit into the few-shot setting. This constitutes more than 90\% of original codes. We then ranked the filtered codes by test/train ratio and select top 50, so that testing samples are available for evaluation.
We also removed some potential common diseases by hand in the process. This would include true rare diseases (e.g. Kaposi's sarcoma) listed in expert labeled rare diseases dictionary \citep{Pavan2017ClinicalPG, NguengangWakap2019EstimatingCP}. We named this dataset \textbf{MIMIC-III-rare50}. The average number of examples per label code (shot) is about 5.

\begin{table*}
\centering
\begin{tabular}{lrrrrrrr}\toprule
\multirow{2}{*}{Model} &\multicolumn{2}{c}{AUC} &\multicolumn{2}{c}{F1} &Precision &\multirow{2}{*}{Best epoch out of 20} \\\cmidrule{2-6}
&Macro &Micro &Macro &Micro &P@5 & \\\midrule
MultiResCNN &89.30 &92.04 &59.29 &66.24 &61.56 &18 \\
MSATT-KG* &91.40 &93.60 &63.80 &68.40 &64.40 &- \\
JointLAAT &92.36 &94.24 &66.95 &70.84 &66.36 &10 \\
MSMN &92.50 &94.39 &67.64 &71.78 &67.23 &15 \\
\hline
KEPTLongformer &\textbf{92.63} &\textbf{94.76} &\textbf{68.91} &\textbf{72.85} &\textbf{67.26} &\textbf{4} \\
 \hspace{3mm} w/o HSAP &92.33 &94.31 &67.95 &71.92 &67.18 &5 \\ 
 \hspace{3mm} w/o HSAP \& Prompt &90.54 &93.18 &58.61 &67.22 &64.38 &17 \\
ClinicalBERT &81.94 &85.65 &43.61 &51.62 &52.59 &15 \\
\bottomrule
\end{tabular}
\caption{Results on the MIMIC-III-50 test set, compared between KEPTLongformer and baselines (top), KEPTLongformer and ablations (down). * represents result collected from paper because no code is avail.}\label{tab:result_common}
\end{table*}

\subsection{Implementation Details}
For medical domain knowledge graph, we used UMLS 2021AA, containing 4.4 million entities.
When mapping entity to its description, we preferred ICD description. If it is not found, then we used UMLS description.
When recreating previous baselines, we used the same hyperparameter setting as mentioned in their published work. We removed R-Drop in \citep{Yuan2022CodeSD} and used plain cross-entropy loss only for a fair comparison among all baselines. 
Code descriptions in prompt use longformer global attention unless otherwise specified.
Our full hyperparameter and config setting using wandb is provided in github.
Self-alignment Pretraining took about 48 hours with 1 NVIDIA V100 GPU. Fine-tuning took about 10 hours with 2 NVIDIA A100 40GB memory GPUs on MIMIC-III-50, and 0.5 hours on MIMIC-III-rare50. During testing, we used dev set to select best threshold for F1 score. Similar to BERT, no hyper-parameters were further searched on the dev set with our longformer. We evaluated with 5 different random seeds for each model and report the median test results across these seeds unless otherwise specified.

\subsection{Baselines}
\noindent \textbf{MultiResCNN} \citep{Li2020ICDCF}
encode free text with Multi-Filter Residual CNN, and applied label code attention mechanism to enable each ICD code to attend different parts of the document.

\noindent \textbf{MSATT-KG} \citep{Xie2019EHRCW}
apply multi-scale attention and graph neural network to capture potential relations between codes, without any changes in the training objectives.

\noindent \textbf{JointLAAT} \citep{Vu2020ALA}
propose a hierarchical joint learning with training objectives to predict both ICD code and its parent ICD code in the hierarchy graph.

\noindent \textbf{MSMN} \citep{Yuan2022CodeSD}
use synonyms with adapted multi-head attention, which achieved SOTA performance on MIMIC-III-50 task.

\subsection{Results}
Results show that our longformer with knowledge pretrained prompt (KEPTLongformer) outperforms the previous state-of-art model MSMN (top of Table \ref{tab:result_common} and Table \ref{tab:result_rare}).
For the common disease code assignment (MIMIC-III-50) task,
our KEPTLongformer achieves macro AUC of 92.63 (+0.13), micro AUC of 94.76 (+0.36), macro F1 of 68.91 (+1.27), and micro F1 of 72.85 (+1.07). Number in parentheses shows the improvements compared to MSMN. For the rare disease code assignment (MIMIC-III-rare50) task,
our KEPTLongformer achieves macro AUC of 82.70 (+7.39), micro AUC of 83.28 (+7.11), macro F1 of 30.44 (+13.39), micro F1 of 32.63 (+15.44). 
We notice that the improvements on rare disease codes are much higher than improvements on common disease code, indicating the strong advantage of our KEPTLongformer for few-shot settings. In contrast to previous work that leads to improvements on rare disease codes but worse results on frequent ones \citep{rios-kavuluru-2018-shot}, our approach shows improvements on both tasks. We finally applied our KEPTLongformer to MIMIC-III-full.
Table \ref{tab:result_full} shows that reranker with KEPTLongformer outperforms previous SOTA MSMN in F1 marco from 10.3 to 11.8 by +1.5 (95\%CI +0.93 to +1.99, P<0.001) and F1 micro from 58.2 to 59.9 by +1.6 (95\%CI +0.95 to +2.33, P<0.001). 

\begin{table*}
\centering
\begin{tabular}{l|l|rrrrrr}\toprule
\multirow{2}{*}{Model} &Setting &\multicolumn{2}{c}{AUC} &\multicolumn{2}{c}{F1} &\multirow{2}{*}{\# Train Param} \\\cmidrule{3-6}
&(trained from) &Macro &Micro &Macro &Micro & \\\midrule
\multirow{3}{*}{MSMN} & \textit{Pretrained}  &75.3 &76.2 &17.1 &17.2 &16.4M \\
&\textit{Finetuned} &58.2 &44.0 &3.3 &4.2 &16.4M \\
&\textit{Zero shot} &52.3 &48.9 &3.5 &4.0 &0 \\
\hline
\multirow{7}{*}{KEPTLongformer} &\textit{Pretrained} &81.4 &82.3 &25.8 &30.9 &119.4M \\
&\textit{Finetuned} &\textbf{82.7} &\textbf{83.3} &\textbf{30.4} &\textbf{32.6} &119.4M \\
& \hspace{3mm} w/o \textit{HSAP} &80.2 &82.2 &24.3 &29.9 &119.4M \\
& \hspace{3mm} w/ \textit{LM only} &75.0 &76.9 &15.2 &16.9&0.6M \\
& \hspace{3mm} w/ \textit{LM} \& \textit{Last} &77.6 &78.4 &17.3 &23.4 &9.4M \\
& \hspace{3mm} w/ \textit{LM} \& \textit{First}  &79.0 &81.5 &23.5 &29.6 &9.4M \\
&\textit{Zero shot} &74.9 &76.5 &15.2 &16.7 &0 \\ 
\bottomrule
\end{tabular}
\caption{Results on the MIMIC-III-rare50 test set compared between MSMN (previous SOTA on MIMIC-III-50) and our final model KEPTLongformer, where \textit{Pretrained}: model is trained from previous pretraining checkpoint, \textit{Finetuned}: model is trained from best checkpoint after finetuned from MIMIC-III-50, \textit{HSAP}: Hierarchical Self-Alignment Pretraining. We also explore training partial model including: parameters of \textit{LM} head, \textit{Last} self-attention layer, \textit{First} self-attention layer as ablation study. \textit{Zero shot}: No training on rare, directly inference using finetuned model from MIMIC-III-50.}\label{tab:result_rare}
\end{table*}

\subsection{Discussion}


Our final KEPTLongformer model could be interpreted as a hybrid of 3 closely interrelated components:
longformer, prompt based fine-tuning, and knowledge injected pretraining. Here we provide an ablation study on each part.

\noindent \textbf{Longformer vs. BERT}.
Increasing max token limit is important under clinical note analysis task, because most clinical notes are long documents with an average of 3000 tokens in MIMIC-III discharge summaries.
Due to the high number of tokens in a medical note, it is essential to encode as many tokens as possible before downstream analysis. However, BERT based LM, which could only encode a few sentences, is known to be ineffective for long documents \citep{Beltagy2020LongformerTL}. To test the effect of max token limit in auto ICD coding task, we compare the performance between Clinical Longformer with max limit of 8,192 tokens and ClinicalBert with max limit of 512 tokens. As shown in Table \ref{tab:result_common}, Clinical Longformer (KEPTLongformer without HSAP \& Prompt) substantially outperforms ClinicalBERT in AUC from 7.5 to 8.6 and F1 from 14.9 to 15.6. 
Other previous methods (e.g. MultiResCNN) use non-pretrained LSTM or CNN with max limit of 8192 tokens.
We also observe that these previous methods outperform ClinicalBERT, indicating the importance of max token limit over LM in auto ICD coding task. This finding correlates to previous LM researches \cite{zhang-etal-2020-bert, pascual-etal-2021-towards, Biswas2021TransICDTB} which only uses longformer/BigBird \citep{Michalopoulos2022ICDBigBirdAC} or hierarchical BERT \citep{Ji2021DoesTM, Dai2022RevisitingTM} of 4096 max token limit, and our method with max limit of 8192 tokens could alleviate the issues mentioned by them.

\noindent \textbf{Prompt based fine-tuning as early fusion}.
In order to test the effect of prompt based fine-tuning as its own, we further compare Longformer trained with prompt based fine-tuning with longformer trained with original fine-tuning on MIMIC-III-50. As shown in Table \ref{tab:result_common}, prompt based fine-tuning (KEPTLongformer w/o HSAP) improves AUC and F1, and converges faster to achieve best F1 score from epoch 17 to epoch 5. 
Our prompt based fine-tuned longformer also slightly outperforms MultiResCNN, and other baselines such as MSATT-KG and JointLAAT that uses structured knowledge as addition resources. 
Under few-shot setting, our prompt based fine-tuning significantly increase AUC and F1 score compared to traditional fine-tuning as shown in Table \ref{tab:result_rare}. This finding supports previous research \citep{Taylor2022ClinicalPL} that shows prompt based fine-tuning outperforms traditional fine-tuning in many few-shot clinical tasks such as length of stay and mortality prediction.
Compared to recent models on auto ICD coding, our prompt based model could be seen as an early \textcolor{CarnationPink}{fusion} of label code description and input note text. 
Instead of fusing label description representations and note text representations after encoder with label attention \citep{zhou-etal-2021-automatic, Dong2021ExplainableAC, Yuan2022CodeSD}, we fuse the two starting from first layer within the encoder with cross attention. Such similar early fusion method has shown to be effective in combining information from knowledge graph and information from text in question answering over knowledge base facts \citep{das-etal-2017-question} and open domain question answering \citep{sun-etal-2018-open}.

\noindent \textbf{Hierarchical self alignment pretraining (HSAP) improves multi-label classification with label domain knowledge}.
In order to test the effect of HSAP as its own, we further compare 
Longformer with HSAP  (KEPTLongformer) and without HSAP (w/o HSAP). HSAP improves 0.45 on micro AUC and 1.09 on micro F1 in dataset MIMIC-III-50, and 1.1 on micro AUC and 2.7 on micro F1 in dataset MIMIC-III-rare50. Thus we showed that our contrastive learning in label space is more effective in the tasks with limited labeled data, which supports similar finding in text classification \citep{Qian2022ContrastiveLF}.
We also observe that HSAP could reduce false negative predictions which mistakenly predict their siblings.
Out of 78 false negative predictions on code 285.1, 2 predict sibling code 285.9 with HSAP.
In contrast, out of 89 false negative predictions on code 285.1, 15 predict sibling code 285.9 without HSAP. 
HSAP reduces false negative predictions on 285.1 caused by sibling 285.9 from 15 to 2. HSAP works as a good polish to further improve the coding accuracy by injecting domain knowledge into language model. 

\noindent \textbf{Parameter efficiency on few-shot learning.}
One could argue that accuracy improvements come from more number of parameters during training. Our KEPTLongformer is finetuned with 7 times more trainable parameters compared to baseline MSMN. To counter such argument, we also fine-tune our KEPTLongformer with limited parameters while keeping most parameters fixed. Specifically, we considered the following 4 settings: a) tuning LM head and first encoder layer, b) tuning LM head and last encoder layer, c) tuning LM head only, d) tuning no parameter as zero-shot. 
Compared to MSMN, settings a, b, c, d improve micro AUC by +5.4, +2.2, +0.7, +0.3 and micro F1 by +12.4, +6.1, -0.3, -0.5 respectively, as shown in Table \ref{tab:result_rare}.
Setting a and b with 9.4 million trainable parameters significantly outperforms MSMN with 16.4 million trainable parameters. Setting c and d with almost no trainable parameters shows competitive results compared to MSMN.
We also observe that training first layer outperforms training last layer, this could also be an evidence to support the advantage of early fusion for few-shot learning.

\begin{table}
\centering
\begin{tabular}{l|rrr}\toprule
Metric &MSMN &Reranker \\\midrule
F1 Mac &10.3(0.3) &11.8(0.4) \\
F1 Mic &58.2(0.4) &59.9(0.5) \\
P@8 &74.9(0.3) &77.1(0.3) \\
R@8 &39.2(0.4) &40.7(0.1) \\
P@15 &59.5(0.1) &61.5(0.2) \\
R@15 &55.7(0.1) &57.4(0.2) \\
\bottomrule
\end{tabular}
\caption{Results on the MIMIC-III-full compared between previous SOTA MSMN and our final model KEPTLongformer reranker. mean(st.dev.) are reported with 5 different random seeds.}
\label{tab:result_full}
\end{table}

\section{Conclusions}
In this paper, we investigate pretrained clinical language model on auto ICD coding task for both common and rare disease, the latter of which has received limited attention in the past. 
Built on recent advances in contrastive learning, entity representation training and prompt based fine-tuning, our KEPTLongformer easily achieves a competitive performance over state of the art system in common code assignment, and significantly outperforms baseline model in rare code assignment task. Finally, our novel Hierarchical Self-Alignment Pretrain could be easily applied to other multi-label classification problems such as tumor detection using other ontology such as OncoTree.


\section{Limitations}
\label{sec:Limitations}

Our work is limited to auto ICD coding task with 50 label codes including MIMIC-III-50 or MIMIC-III-rare50, and could not be directly applied to ICD coding task MIMIC-III-full with 8,692 labels in practice due to memory constraint. Using our KEPTLongformer would create at least 26,076 tokens and 8,692 [MASK] in a single prompt, which easily explodes the max token limit of a longformer and GPU memory.
A more memory efficient method for auto ICD coding could be explored for future work.


Our clinical knowledge pretrained KEPTLongformer is only tested on auto ICD coding task, but such pretrained language model could be easily applied to other clinical NLP applications such as clinical entity linking or clinical question answering tasks. 
We also only use part of UMLS knowledge graph, including hierarchy, synonym, and abbreviation. Other knowledge including disease co-occurrence, disease-symptom, disease-lab relations and others could also potentially useful for auto ICD coding task. 

\section*{Acknowledgements}
We are grateful to the UMass BioNLP and MLFL group for many helpful discussions and related talks which inspired this work. We would also like to thank the anonymous reviewers for their insightful feedback. Research reported in this study was supported by the National Science Foundation under award 2124126. The work was also in part supported by the National Institutions of Health R01DA045816 and R01MH125027. The content is solely the responsibility of the authors and does not necessarily represent the official views of the National Science Foundation and National Institutes of Health.

\bibliography{anthology,custom}
\bibliographystyle{acl_natbib}

\newpage
\appendix
\renewcommand\thefigure{\thesection.\arabic{figure}}  
\renewcommand\thetable{\thesection.\arabic{table}}

\section{Appendix}
\setcounter{figure}{0}
\setcounter{table}{0}

\begin{table}[ht]
\centering
\begin{tabular}{l}\toprule
Section header \\\midrule
chief complaint: \\
procedure: \\
history of present illness: \\
past medical history: \\
brief hospital course: \\
discharge diagnosis: \\
discharge condition: \\
\bottomrule
\end{tabular}
\caption{A list of section header names used to truncate if document token length > 8192.}\label{tab:header}
\end{table}

\begin{table}[ht]
\centering
\begin{tabular}{lcc}
\hline
\textbf{MIMIC-III-full} & \textbf{train} & \textbf{test}\\
\hline
\#Doc. & 47,723 & 3,372 \\
Avg \#words per Doc. & 1,504 & 1,818 \\
Avg \#tokens per Doc. & 2,479 & 3,071 \\
\%Doc where \#tokens < 512 & 1.1 & 0.1  \\
\%Doc where \#tokens < 4096 & 89.3 & 82.3  \\
\%Doc where \#tokens < 8192 & 99.6 & 99.4 \\
Avg \#codes per Doc. & 15.6 & 17.9 \\ 
\hline
\textbf{MIMIC-III-50} & \textbf{train} & \textbf{test}\\
\hline
\#Doc. & 8,066 & 1,729 \\
Avg \#tokens per Doc. & 3,008 & 3,665 \\
\%Doc where \#tokens < 512 & 0.5 & 0.1  \\
\%Doc where \#tokens < 4096 & 80.1 & 67.7  \\
\%Doc where \#tokens < 8192 & 97.9 & 98.7 \\
Avg \#codes per Doc. & 5.7 & 6.0 \\ 
\hline
\textbf{MIMIC-III-rare50} & \textbf{train} & \textbf{test}\\
\hline
\#Doc. & 249 & 142 \\
Avg \#tokens per Doc. & 3,462 & 4,131 \\
\%Doc where \#tokens < 512 & 0.1 & 0.1  \\
\%Doc where \#tokens < 4096 & 71.1 & 55.6  \\
\%Doc where \#tokens < 8192 & 96.8 & 96.5\\
Avg \#codes per Doc. & 1.0 & 1.0 \\ 
\hline
\end{tabular}
\caption{Statistics of MIMIC-III dataset under full codes settings (MIMIC-III-full), 50 common codes settings (MIMIC-III-50), and 50 rare codes settings (MIMIC-III-rare50).} \label{tab:data}
\end{table}

\begin{table*}
\centering
\begin{tabular}{lrrrrr}\toprule
Model &\multicolumn{2}{c}{AUC} &\multicolumn{2}{c}{F1} \\\cmidrule{1-5}
KEPTLongformer &Macro &Micro &Macro &Micro \\\midrule
InfoNCE &92.48 &94.32 &68.38 &72.01 \\
Hierarchical contrastive loss &\textbf{92.63} &\textbf{94.76} &\textbf{68.91} &\textbf{72.85} \\
\bottomrule
\end{tabular}
\caption{Results on the MIMIC-III-50, compared between KEPTLongformer using hierarchical contrastive loss in this work, compared to InfoNCE which is used in KRISSBERT.}\label{tab:result_contrastive}
\end{table*}

\begin{table*}
\centering
\begin{tabular}{lrrrrrrr}\toprule
Model &\multicolumn{2}{c}{AUC} &\multicolumn{2}{c}{F1} &Precision &\multirow{2}{*}{Best epoch out of 20} \\\cmidrule{2-6}
KEPTLongformer&Macro &Micro &Macro &Micro &P@5 & \\\midrule
3 layers &88.65 &91.97 &57.75 &64.67 &61.87 &18 \\
6 layers &91.99 &94.41 &66.94 &71.33 &64.67 &7 \\
12 layers &\textbf{92.63} &\textbf{94.76} &\textbf{68.91} &\textbf{72.85} &\textbf{67.26} &\textbf{4} \\
\bottomrule
\end{tabular}
\caption{Results on the MIMIC-III-50, compared between KEPTLongformer using different number of layers.}\label{tab:result_layer}
\end{table*}

\begin{table*}
\centering
\begin{tabular}{lrrrrrrr}\toprule
Model &\multicolumn{2}{c}{AUC} &\multicolumn{2}{c}{F1} &Precision &\multirow{2}{*}{Memory} \\\cmidrule{1-6}
KEPTLongformer &Macro &Micro &Macro &Micro &P@5 & \\\midrule
global stride = 1 &\textbf{92.63} &\textbf{94.76} &\textbf{68.91} &\textbf{72.85} &\textbf{67.26} &34G \\
global stride = 3 &92.49 &94.55 &68.43 &72.30 &67.23 &27G \\
global stride = 5 &92.24 &94.46 &68.09 &72.17 &66.93 &25G \\
global stride = 10 &92.02 &94.24 &66.98 &71.14 &65.45 &\textbf{23G} \\
\bottomrule
\end{tabular}
\caption{Results on the MIMIC-III-50, compared between KEPTLongformer using different number of global attentions. In prompt code descriptions, we set every $global\_stride$ number of tokens as (longformer) global attention tokens. For example, global stride = 1 means each token in prompt code descriptions is global attention. $Memory$ is the required GPU Memory when $per\_device\_train\_batch\_size=1$.}
\label{tab:result_gstride}
\end{table*}

\end{document}